\documentclass[11pt,a4paper]{article}
\usepackage{times,latexsym}
\usepackage{url}
\usepackage[T1]{fontenc}

\usepackage{amsmath}
\usepackage{bbding}
\usepackage{array}
\usepackage[table]{xcolor}
\usepackage{amssymb}
\usepackage{tikz}
\usepackage{forest}
\usetikzlibrary{mindmap, shadows}
\usetikzlibrary{arrows,shapes,positioning,shadows,trees,arrows.meta}

\tikzset{
  basic/.style  = {draw, rounded corners=2pt, font=\small, thin, align=center, text width=8cm, drop shadow, rectangle}
}

\usepackage[acceptedWithA]{tacl2021v1}

\usepackage{xspace,mfirstuc,tabulary}

\newif\iftaclinstructions
\taclinstructionstrue 
\iftaclinstructions
\renewcommand{\confidential}{}
\renewcommand{\anonsubtext}{(No author info supplied here, for consistency with
TACL-submission anonymization requirements)}
\newcommand{\instr}
\fi

\iftaclpubformat 

\else

\fi

\title{Attribution Techniques for Mitigating Hallucinated Information in RAG Systems: A Survey}

\author{
  Yuqing Zhao \quad Ziyao Liu \quad Yongsen Zheng\textdagger \quad Kwok-Yan Lam \\
  Nanyang Technological University, Singapore \\
  \texttt{ZHAO0522@e.ntu.edu.sg, liuziyao, yongsen.zheng, kwokyan.lam@ntu.edu.sg}
}
\date{}

\begin{document}
\maketitle

\renewcommand{\thefootnote}{}
\footnotetext{\textsuperscript{\textnormal{\textdagger}}Corresponding author.
}

\begin{abstract}
Large Language Models (LLMs)-based question answering (QA) systems play a critical role in modern AI, demonstrating strong performance across various tasks. However, LLM-generated responses often suffer from hallucinations, unfaithful statements lacking reliable references. Retrieval-Augmented Generation (RAG) frameworks enhance LLM responses by incorporating external references but also introduce new forms of hallucination due to complex interactions between the retriever and generator. To address these challenges, researchers have explored attribution-based techniques that ensure responses are verifiably supported by retrieved content. Despite progress, a unified pipeline for these techniques, along with a clear taxonomy and systematic comparison of their strengths and weaknesses, remains lacking. A well-defined taxonomy is essential for identifying specific failure modes within RAG systems, while comparative analysis helps practitioners choose appropriate solutions based on hallucination types and application context. This survey investigates how attribution-based techniques are used within RAG systems to mitigate hallucinations and addresses the gap by: (i) outlining a taxonomy of hallucination types in RAG systems, (ii) presenting a unified pipeline for attribution techniques, (iii) reviewing techniques based on the hallucinations they target, and (iv) discussing strengths and weaknesses with practical guidelines. This work offers insights for future research and practical use of attribution techniques in RAG systems.
\end{abstract}

\section{Introduction}
Currently, Artificial Intelligence (AI) is widely applied across various applications, including question answering (QA) \cite{HutCRS, HyFairCRS, HyberCRS, HiCore}, multi-agent systems \cite{CausalGPT, agent1}, dialogue systems \cite{HyCoRec, CIREC, CIPL, MarkovTree}, machine unlearning systems \cite{LLMRisk,MUSURVEY}, and recommender systems \cite{Music_Rec, GCFM, FacetCRS, DM}. The emergence of Large Language Models (LLMs) has further accelerated the development of these domains. Trained on extensive corpora of textual data, these models demonstrate remarkable generalization abilities, allowing them to generate fluent, coherent, and contextually relevant text. However, despite their impressive performance, LLMs face a significant challenge: hallucination. Hallucination refers to the generation of responses that are factually incorrect, unsupported by retrieved references, or misaligned with user queries. Such inaccuracies can severely undermine the reliability and safety of LLM-based systems, particularly in high-stakes domains such as healthcare, law, and education.

To address this issue, Retrieval-Augmented Generation (RAG) systems have emerged as a promising solution. RAG integrates Large Language Models (LLMs) with external knowledge sources, thereby enhancing factual grounding through the retrieval of relevant references that inform the generation process. This architecture not only improves the faithfulness of generated outputs but also enhances interpretability, allowing users to trace generated responses back to their supporting content. However, while RAG effectively mitigates certain types of hallucination inherent in LLMs, it also introduces new challenges. The retrieval process may yield outdated, irrelevant, or unverifiable information, which can mislead the generator and result in new forms of hallucination \cite{11131222}. Specifically, errors or mismatches between the retriever and generator—such as the retrieval of irrelevant references—can propagate into the final responses, complicating the overall reliability of the system.

Recent studies have investigated attribution-based techniques to mitigate hallucination in RAG systems. These methods aim to enhance the faithfulness of generated responses by ensuring they are accurately supported by verifiable references. Key approaches include refining input queries to improve retrieval accuracy, filtering or reranking retrieved documents for relevance, crafting prompt templates that encourage grounded generation, and applying post-hoc corrections to the generated responses. Despite these advancements, a unified understanding of which techniques effectively address specific types of hallucination remains elusive, particularly in the context of RAG-specific failure points that arise from the interaction between the retriever and generator.

\textbf{Comparison with related surveys.}
While several recent surveys offer valuable overviews of hallucination in LLMs, they either address LLMs in general or lack specificity in correlating techniques with the causes of hallucination in Retrieval-Augmented Generation (RAG) systems. For instance, surveys such as \cite{zhang2023siren, huang2025survey, ye2023cognitive, tonmoy2024comprehensive} analyze hallucination mechanisms in LLMs broadly but do not consider the unique error sources introduced by RAG architectures. Conversely, others, like \cite{li2023survey, gupta2024comprehensive, zhang2023siren, gao2023retrieval}, examine attribution techniques or the evolution of RAG but fail to provide a taxonomy that links specific types of hallucination to appropriate mitigation strategies. This survey addresses this gap by explicitly focusing on hallucinations within RAG systems and organizing attribution methods based on the interactions between the retriever and the generator.

\textbf{Summary of contributions.}

1) We propose a comprehensive taxonomy of hallucination types that specifically arise from the interaction between the retriever and the generator in RAG systems. 

2) We introduce a unified pipeline for attribution-based hallucination mitigation, composed of four modular components—Query Refining (T1), Reference Identification (T2), Prompt Engineering (T3), and Response Correction (T4)—that require no changes to the model architecture.

3) We systematically map each module to the hallucination types they best mitigate, offering practical guidance on selecting appropriate techniques based on the source of error. 

4) We provide a comparative analysis of attribution methods in terms of effectiveness, computational efficiency, and application domains, and discuss usage strategies, implementation trade-offs, and open challenges in real-world settings.

\section{A Unified Pipeline of LLM Attribution}

\begin{figure}[t]
\centering
  \includegraphics[scale=0.24]{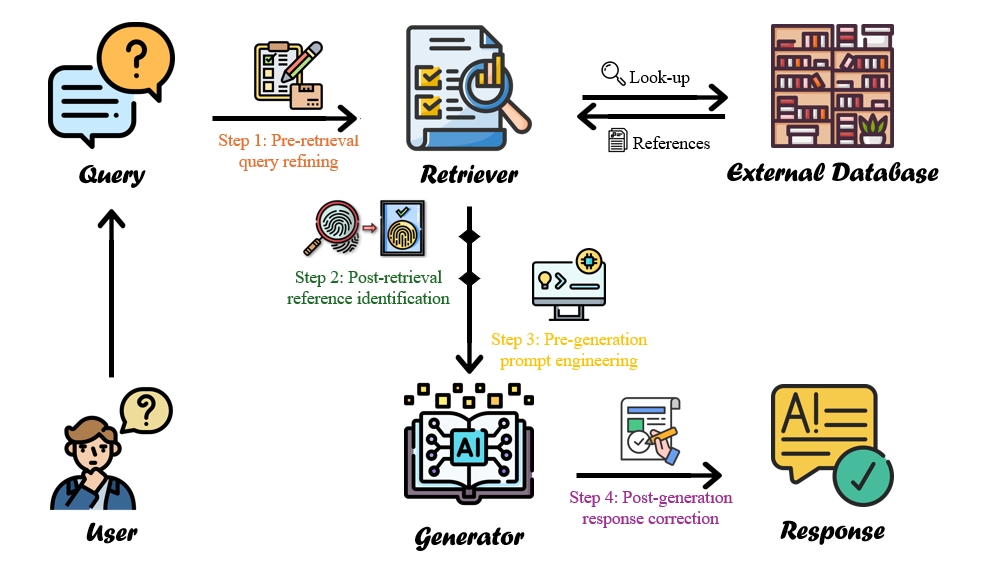}
  \caption{A Unified Pipeline of LLM Attribution}
  \label{fig:flow}
\end{figure}

The LLM attribution pipeline integrates LLMs with external knowledge sources to generate responses supported by credible references. When a user submits a query \( q \), the retriever identifies relevant references \( R' \). The generator synthesizes a response \( A \) using \( q \) and \( R' \) as \( A = \mathcal{M}(q, \{r\}), \forall r \in R' \). The pipeline consists of four modules (as shown in Figure \ref{fig:flow}): Pre-retrieval Query Refining (T1), Post-retrieval Reference Identification (T2), Pre-generation Prompt Engineering (T3), and Post-generation Response Correction (T4). These modules enhance efficiency, accuracy, and user experience without altering the model architecture or requiring fine-tuning.

\textbf{Pre-retrieval Query Refining (T1)}: Refines the input query \( q \) to \( q' \) to optimize retrieval. The transformation is defined as \( q' = \mathcal{T}(q) \).

\textbf{Post-retrieval Reference Identification (T2)}: Identifies the most relevant references from \( R' \), producing \( R_{\text{final}} \subseteq R' \) through the transformation \( R_{\text{final}} = \mathcal{V}(R') \).

\textbf{Pre-generation Prompt Engineering (T3)}: Creates a structured prompt \( P \) by combining \( q' \) and \( R_{\text{final}} \) to enhance the generator's input, defined as \( P = \mathcal{S}(q', R_{\text{final}}) \).

\textbf{Post-generation Response Correction (T4)}: Reviews the generated response \( A_{\text{gen}} \) for accuracy and coherence, adjusting it as needed with the function \( A_{\text{corrected}} = \mathcal{C}(q', A_{\text{gen}}, R_{\text{final}}) \).

\begin{figure*}
\begin{tikzpicture}[
  level 1/.style={sibling distance=27mm},
  level distance=13.5mm,
  edge from parent/.style={->,draw},
  >=latex]

\node[basic, font=\normalsize, fill=gray!20] (parent) {Taxonomy of Hallucination and \\ Mitigation via Attribution Techniques}
  child [edge from parent/.style={draw=none}]{node[basic, text width=22mm, fill=red!40] (c1) {Overconfidence  \\ Hallucination}}
  child {node[basic, text width=21mm, fill=blue!40] (c2) 
  {Outdatedness  \\ Hallucination}}
  child {node[basic, text width=21mm, fill=green!40] (c3) {Unverifiability \\ Hallucination}}
  child {node[basic, text width=21mm, fill=cyan!40] (c4) 
  {Instruction \\ Deviation}}
  child {node[basic, text width=21mm, fill=orange!40] (c5) 
  {Context \\ Inconsistency}}
  child [edge from parent/.style={draw=none}]{node[basic, text width=21mm, fill=yellow!40] (c6) {Reasoning \\ Deficiency}};

\begin{scope}
\node[basic, text width=21mm,node distance=11mm,  below of = c1, fill=red!20, xshift=5pt] (c11) {Pre-retrieval \\ query refining};
\node[basic, text width=21mm,node distance=57mm, below of = c11, fill=red!20] (c12) {Post-retrieval \\ reference identification};

\node[basic, text width=21mm, node distance=11mm, below of = c2, fill=blue!20, xshift=5pt] (c21) {Pre-retrieval \\ query refining};
\node[basic, text width=21mm,node distance=38mm, below of = c21, fill=blue!20] (c22) {Post-retrieval \\ reference identification};

\node[basic, text width=21mm, node distance=13mm, below of = c3, fill=green!20, xshift=5pt] (c31) {Post-retrieval \\ reference identification};
\node[basic, text width=21mm, node distance=83.5mm, below of = c31, fill=green!20] (c32) {Post-generation \\ response correction
};

\node[basic, text width=21mm, node distance=11.5mm, below of = c4, fill=cyan!20, xshift=5pt] (c41) {Pre-retrieval \\ query refining};
\node[basic, text width=21mm, node distance=130.5mm, below of = c41, fill=cyan!20] (c42) {Post-retrieval \\ reference identification};
\node[basic, text width=21mm, node distance=51.5mm, below of = c42, fill=cyan!20] (c43) {Pre-generation \\prompt engineering};

\node[basic, text width=21mm, node distance=14mm, below of = c5, fill=orange!20, xshift=5pt] (c51) {Pre-generation \\ prompt engineering};
\node[basic, text width=21mm, node distance=69mm, below of = c51, fill=orange!20] (c52) {Post-generation \\ response correction};

\node[basic, text width=21mm, node distance=14mm, below of = c6, fill=yellow!20, xshift=5pt] (c61) {Pre-generation \\prompt engineering};
\node[basic, text width=21mm, node distance=43mm, below of = c61, fill=yellow!20] (c62) {Post-generation \\ response correction};

\node [basic, align=left, node distance=28mm, font=\footnotesize, text width=21mm, below of = c11, xshift=0pt, fill=white] (c111) {SAPLMA \cite{SAPLMA}; IFL \cite{IFL}; SelfCheckGPT \cite{Selfcheckgpt}; RLKF \cite{RLKF}};
\node [basic, align=left, node distance=14mm, text width=21mm, font=\footnotesize, below of = c12, xshift=0, fill=white] (c121) {VUC \cite{VUC}};

\node [basic, align=left, node distance=18.5mm, text width=21mm, font=\footnotesize, below of = c21, xshift=0, fill=white] (c211) {WebCPM \cite{Webcpm}; SmartBook \cite{Smartbook}};
\node [basic, align=left, node distance=47mm, text width=21mm, font=\footnotesize, below of = c22, xshift=0, fill=white] (c221) {LLM-Augmenter \cite{LLM-Augmenter}; DPR \cite{DPR}; WebGPT \cite{Webgpt}; WebBrain \cite{Webbrain}; RAG \cite{RAG}; CoDA \cite{CoDA}; REALM \cite{REALM}; FDP \cite{FDP}};

\node [basic, align=left, node distance=41mm, text width=21mm, font=\footnotesize, below of = c31, fill=white] (c311) {Replug \cite{Replug}; AAR \cite{AAR}; Self-rag \cite{Self-rag}; CoTAR \cite{CoTAR}; C-RAG \cite{kang2024cragcertifiedgenerationrisks}; PRCA \cite{PRCA}; Recomp \cite{Recomp}; TOC \cite{TOC}};
\node [basic, align=left, node distance=67mm, text width=21mm, font=\footnotesize, below of = c32, xshift=0, fill=white] (c321) {LaMDA \cite{Lamda}; Sparrow \cite{Sparrow};  GopherCite \cite{GopherCite}; TWEAK \cite{TWEAK}; Atlas \cite{Atlas}; Fine-grained RLHF \cite{Fine-grainedRLHF}; SELF-REFINE \cite{Self-refine}; RARR \cite{RARR}; AGREE \cite{agree}; CaLM \cite{CaLM}; ITRG \cite{ITRG}; ITER-RETGEN \cite{ITER-RETGEN}};

\node [basic, align=left, node distance=65mm, text width=21mm, font=\footnotesize, below of = c41, xshift=0, fill=white] (c411) {MixAlign \cite{MixAlign}; 1-PAGER \cite{1-pager}; CCV \cite{CCV}; Blueprint \cite{Blueprint}; DSP \cite{DSP}; TOC \cite{TOC}; Step-BackPrompting \cite{Step-BackPrompting}; VTG \cite{VTG}; Query2doc \cite{Query2doc}; In-Context RALM \cite{In-contextRALM}; HyDE \cite{HyDE}; RRR \cite{RRR}};
\node [basic, align=left, node distance=26mm, text width=21mm, font=\footnotesize, below of = c42, xshift=0, fill=white] (c421) {Self-RAG \cite{Self-rag}; RARR \cite{RARR}; Llatrieval \cite{Llatrieval}; QLM \cite{QLM}};
\node [basic, align=left, node distance=17mm, text width=21mm, font=\footnotesize, below of = c43, xshift=0, fill=white] (c431) {UPRISE \cite{Uprise}};

\node [basic, align=left, node distance=34mm, text width=21mm, font=\footnotesize, below of = c51, xshift=0, fill=white] (c511) {QUIP \cite{QUIP}; RECITE \cite{RECITE}; Smartbook \cite{Smartbook}; PKG \cite{PKG}; In-Context RALM \cite{In-contextRALM}};
\node [basic, align=left, node distance=32mm, text width=21mm, font=\footnotesize, below of = c52, xshift=0, fill=white] (c521) {SourceCheckup \cite{SourceCheckup}; EFEC \cite{EFEC}; RSEQGA \cite{RSEGQA}; WebCiteS \cite{WebCiteS}};

\node [basic, align=left, node distance=21mm, text width=21mm, font=\footnotesize, below of = c61, xshift=0, fill=white] (c611) {SubgraphRAG \cite{SubgraphRAG}; Self-Reasoning \cite{Self-Reasoning}};
\node [basic, align=left, node distance=28mm, text width=21mm, font=\footnotesize, below of = c62, xshift=0, fill=white] (c621) { CoVe \cite{CoVe}; Searchain \cite{Searchain}; Verify and Edit \cite{Verify-and-edit}};

\end{scope}

\draw[->] (parent.190) --  (c1.25);  
\draw[->] (parent.350) --  (c6.148);   

\foreach \value in {1,2}
  \draw[->] (c1.195) |- (c1\value.west);

\foreach \value in {1,2}
  \draw[->] (c2.195) |- (c2\value.west);

\foreach \value in {1,2}
  \draw[->] (c3.195) |- (c3\value.west);

\foreach \value in {1,2,3}
  \draw[->] (c4.195) |- (c4\value.west);

\foreach \value in {1,2}
  \draw[->] (c5.195) |- (c5\value.west);

\foreach \value in {1,2}
  \draw[->] (c6.195) |- (c6\value.west);
\draw[->] (c11.270) -- (c111.north);
\draw[->] (c12.270) -- (c121.north);

\draw[->] (c21.270) -- (c211.north);
\draw[->] (c22.270) -- (c221.north);

\draw[->] (c31.270) -- (c311.north);
\draw[->] (c32.270) -- (c321.north);

\draw[->] (c41.270) -- (c411.north);
\draw[->] (c42.270) -- (c421.north);
\draw[->] (c43.270) -- (c431.north);

\draw[->] (c51.270) -- (c511.north);
\draw[->] (c52.270) -- (c521.north);

\draw[->] (c61.270) -- (c611.north);
\draw[->] (c62.270) -- (c621.north);
\end{tikzpicture}

\caption{Taxonomy of Attribution for Mitigating six types of LLM Hallucinations.}
\label{fig:taxonomy}
    
\end{figure*}
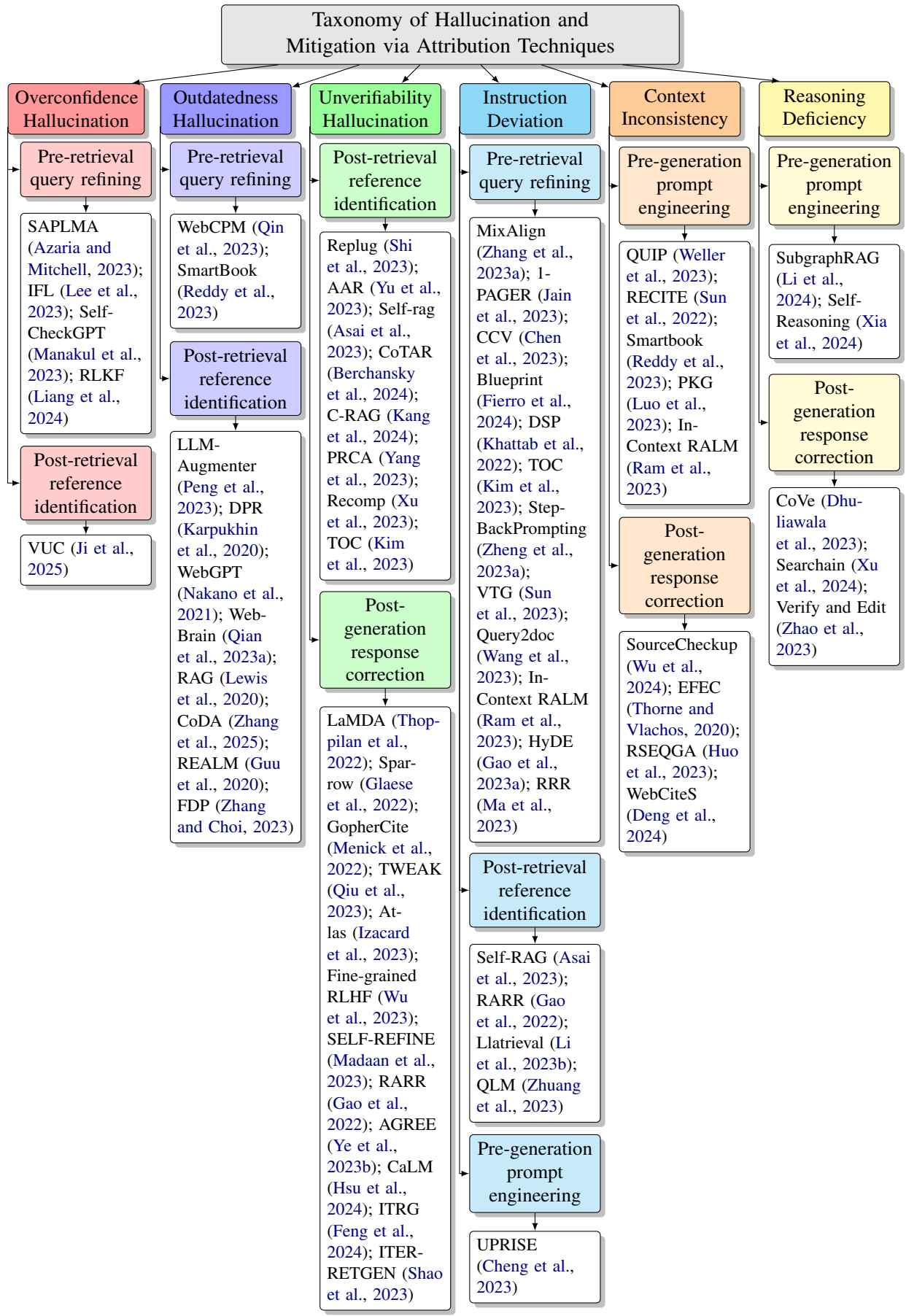

\section{Taxonomy of Attribution for Mitigating Hallucinations} 
Hallucination occurs when an LLM generates responses that are factually incorrect, unsupported, or misaligned with the query or references. In RAG systems, hallucinations can arise from the retrieval process (e.g., outdated or irrelevant references) or the generation process (e.g., faulty reasoning or misinterpretation). This section presents a taxonomy of hallucination types specific to RAG systems, based on the interaction between the retriever and generator, as illustrated in Figure \ref{fig:taxonomy}. This taxonomy provides a framework for mapping attribution-based techniques to specific hallucination types.

\subsection{Regulating Overconfidence Hallucination}
Overconfidence hallucination happens when an LLM presents uncertain or incorrect information with unwarranted certainty. This section reviews two common attribution techniques to mitigate this issue: Pre-generation Prompt Engineering (T3) and Post-generation Response Correction (T4).

\textbf{Pre-generation Prompt Engineering (T3)} 
Overconfidence hallucination occurs when a response expresses excessive certainty about uncertain information. Adding hedging words like "likely" or "possibly" can mitigate this. Pre-generation prompt engineering directly addresses this by incorporating instructions to include hedging words. A notable technique, Verbal Uncertainty Calibration (VUC) \cite{VUC}, strategically adds these words in the prompt to align the generated response's confidence level with that of the retrieved references, measured by a verbal uncertainty feature in the representation space.

\textbf{Post-generation Response Correction (T4)} 
Overconfidence hallucination can stem from over-reliance on specific references and mismatched certainty levels between generated responses and retrieved references. Post-generation Response Correction methods address this by reducing reliance on dominant references \cite{Selfcheckgpt} and adjusting response certainty \cite{IFL, RLKF, SAPLMA}. Some methods critique and correct overconfident outputs. For instance, RLKF \cite{RLKF} uses a reward model to penalize false confidence and reinforce appropriately uncertain responses, guiding the LLM to match its certainty level with its internal knowledge. Similarly, SAPLMA \cite{SAPLMA} employs a classifier to detect confident statements known to be false, flagging and prompting regeneration of such responses. To mitigate reliance on specific references, Iterative Feedback Learning (IFL) \cite{IFL} uses a critic model to assess the correctness and citation quality of initial responses, guiding the generator to include more diverse references through iterative refinement. Instead of evaluating a single response's confidence, SelfCheckGPT \cite{Selfcheckgpt} assesses internal confidence consistency across multiple responses, selecting the one most aligned with retrieved references. This approach helps identify overconfident statements lacking support in alternative responses.

\subsection{Managing Outdatedness Hallucination}
Outdatedness hallucination occurs when an LLM generates responses that were previously correct but are now obsolete. To address this, attribution-based techniques offer cost-effective solutions. This section examines two common approaches: Pre-retrieval Query Refining (T1) and Post-retrieval Reference Identification (T2).

\textbf{Pre-retrieval Query Refining (T1)} Outdatedness hallucination can arise from ambiguous queries or the absence of time references, leading to the retrieval of outdated information. Pre-retrieval query refining methods can mitigate this by enhancing the precision and timeliness of retrieval \cite{Webcpm, Smartbook}. For instance, SmartBook \cite{Smartbook} adds keywords to queries to limit retrieved references to recent sources, such as those from the past two weeks. Similarly, WebCPM \cite{Webcpm} enhances WebGPT through iterative query refinement techniques like synonym substitution and paraphrasing to improve the timeliness of retrieved references.

\textbf{Post-retrieval Reference Identification (T2)} Outdatedness hallucination often stems from misidentifying current references. To tackle this, Post-retrieval Reference Identification methods have been extensively studied \cite{DPR, RAG, LLM-Augmenter, CoDA, REALM, Webbrain, Webgpt, FDP}. Pioneering techniques like Dense Passage Retrieval (DPR) \cite{DPR} assess the timeliness of references based on relevance, frequency, and source reliability, making these retrievers foundational in RAG systems \cite{RAG, Atlas}, where identified references inform final responses (see Figure \ref{fig:flow}). Subsequent works have aimed to enhance retrieval timeliness. LLM-Augmenter \cite{LLM-Augmenter} employs Reinforcement Learning with Human Feedback (RLHF) to refine retrieval strategies for specific dates. CoDA \cite{CoDA} addresses outdatedness by prioritizing timely over popular references, reducing the weight of less relevant candidates. Unlike one-shot enhancements, REALM \cite{REALM} dynamically optimizes retrieval based on database updates. Fact Duration Prediction (FDP) \cite{FDP} introduces a model to evaluate the temporal validity of references. Additionally, WebBrain \cite{Webbrain} maintains a continuously updated database for Wikipedia tasks, while WebGPT \cite{Webgpt} simulates real-time web searches, effectively serving as an up-to-date knowledge base. This approach is now common in LLM chatbot applications like ChatGPT, DeepSeek, and Anthropic.

\subsection{Alleviating Unverifiability Hallucination}
Unverifiability hallucination arises when an LLM generates responses without adequate supporting references, leading to inaccuracies. To counter this, attribution-based methods retrieve evidence from independent databases, gathering information until sufficient references are collected for a verifiable response. This section explores Post-retrieval Reference Identification (T2) and Post-generation Response Correction (T4) as strategies to mitigate unverifiability hallucinations.

\textbf{Post-retrieval Reference Identification (T2)} 
Unverifiability hallucination occurs when an LLM generates content unsupported by retrieved evidence, often due to loosely related references. To address this, Post-retrieval Reference Identification methods ensure that only verifiable and well-aligned references support the final response, using techniques like references reranking \cite{Replug, AAR, Self-rag, kang2024cragcertifiedgenerationrisks} or selecting finer-grained references \cite{CoTAR, PRCA, Recomp, TOC}. A direct approach involves leveraging LLMs to enhance retrieval. Methods such as REPLUG \cite{Replug}, AAR \cite{AAR}, and SELF-RAG \cite{Self-rag} use a separate LLM to re-rank retrieved results, assessing references as fully supported, partially supported, or unsupported to select the most relevant ones. C-RAG \cite{kang2024cragcertifiedgenerationrisks} uses a pretrained transformer model to evaluate reasoning paths and rerank references based on the verifiability score rather than mere similarity. To improve attribution quality, several methods focus on finer-grained references and structured reasoning. For example, CoTAR \cite{CoTAR} prompts the LLM to reason through evidence step by step, enabling multi-level attribution that links specific segments of references to the response. PRCA \cite{PRCA} condenses supportive sentences from retrieved references into concise inputs for the generator. RECOMP \cite{Recomp} identifies supportive references by summarizing them into query-relevant text. Lastly, TOC \cite{TOC} generates clarifying questions using few-shot prompting, organizes them into a tree structure, and validates responses to produce comprehensive long-form outputs.

\textbf{Post-generation Response Correction (T4)} 
Unverifiability hallucination arises when relevant but incomplete references lead the model to generate unverifiable content. Post-generation Response Correction methods aim to (i) correct a single response using higher-quality references \cite{Self-refine, RARR, agree, CaLM, ITRG, ITER-RETGEN} or (ii) generate and rank multiple responses for relevance \cite{TWEAK, GopherCite, Lamda, Fine-grainedRLHF, Sparrow}. Some techniques leverage the LLM or auxiliary verifiers to check reference support. For example, Self-Refine \cite{Self-refine} prompts the LLM to critique its output and retrieve additional evidence if needed. CaLM \cite{CaLM} verifies cited statements, retaining only confirmed parts. AGREE \cite{agree} uses Natural Language Inference to detect unverifiable content, prompting targeted evidence retrieval. RARR \cite{RARR} employs a multi-step process to generate sub-questions for further evidence. ITER-RETGEN \cite{ITER-RETGEN} and ITRG \cite{ITRG} iteratively refine responses by expanding queries. Another strategy involves generating multiple responses and ranking them. TWEAK \cite{TWEAK} assesses supportiveness with a Hypothesis Verification Model, while RLHF-based methods like GopherCite \cite{GopherCite} and Fine-Grained RLHF \cite{Fine-grainedRLHF} score verifiability and provide feedback. LaMDA \cite{Lamda} filters responses for factuality, and Sparrow \cite{Sparrow} applies rule-based evaluations to re-rank or filter out non-compliant responses.

\subsection{Correcting Instruction Deviation}
Instruction deviation hallucination happens when an LLM generates responses that do not follow given instructions, leading to topic shifts, incomplete answers, or inconsistencies. Attribution-based solutions use external guidance to refine queries, identify relevant references, and prompt templates, ensuring alignment with user instructions. This section covers three common techniques for correcting instruction deviation hallucination: Pre-retrieval Query Refining (T1), Post-retrieval Reference Identification (T2), and Pre-generation Prompt Engineering (T3).

\textbf{Pre-retrieval Query Refining (T1)} 
Instruction deviation hallucination occurs when a model's response misaligns with user intent, often due to ambiguous queries. To address this, Pre-retrieval Query Refining methods are employed to clarify user input, ensuring retrieval aligns with the intended task and minimizes instruction-following errors. These methods include (i) keyword-based refinement \cite{MixAlign, 1-pager}, (ii) query decomposition into subqueries \cite{Blueprint, CCV, RRR, Step-BackPrompting, TOC, DSP}, and (iii) applying explicit or learned rules \cite{In-contextRALM, VTG, Query2doc, HyDE, RRR}. Keyword-based refinement methods like MixAlign \cite{MixAlign} and 1-PAGER \cite{1-pager} enhance alignment with user intent. MixAlign uses LLMs to identify vague constraints and prompts users for clarification, while 1-PAGER extracts keywords to iteratively filter irrelevant references, progressively narrowing down to highly relevant information. Query decomposition methods, such as Blueprint \cite{Blueprint}, convert complex queries into structured sub-questions, clarifying expected content. CCV \cite{CCV}, DSP \cite{DSP}, and TOC \cite{TOC} rewrite specific points into multiple sub-questions for targeted retrieval. Step-Back Prompting \cite{Step-BackPrompting} transforms queries into broader questions to guide reasoning and improve accuracy. Rule-based refinement includes In-Context RALM \cite{In-contextRALM}, which adjusts retrieval windows to maintain contextual relevance, and VTG \cite{VTG}, which modifies queries based on semantic relationships with retrieved results. Query2doc \cite{Query2doc} and HyDE \cite{HyDE} generate hypothetical documents to expand queries, while RRR \cite{RRR} enhances user intent expression for improved retrieval.

\textbf{Post-retrieval Reference Identification (T2)} 
Instruction deviation hallucination can occur when retrieved references are topically relevant but misaligned with the user's specific intent. This disconnect may lead to off-topic responses or missed objectives. To this end, Post-retrieval Reference Identification methods are used to evaluate and re-rank retrieved references based on their alignment with the query \cite{Self-rag, RARR, Llatrieval, QLM}. Several methods employ semantic relevance filtering for reference reranking. Self-RAG \cite{Self-rag} uses a relevance estimator that combines semantic similarity (via dense retrieval) and keyword matching (e.g., TF-IDF, BM25) to prioritize relevant references. RARR \cite{RARR} enhances this with an NLI-based model that ranks references based on their semantic support for the query. LLatrieval \cite{Llatrieval} and QLM \cite{QLM} take this further by using an LLM to verify the relevance of retrieved references iteratively, ensuring the final set meets a predefined semantic alignment threshold and aligns closely with user intent.

\textbf{Pre-generation Prompt Engineering (T3)} Pre-generation Prompt Engineering techniques mitigate instruction deviation hallucination by creating prompt templates that clearly define the instructional objective before generation \cite{Uprise}. A notable method is UPRISE \cite{Uprise}, which generates a pool of prompt templates from retrieved references and selects the most aligned one for guiding the generator. Unlike static templates, UPRISE uses an additional retriever to fetch optimal prompts, ensuring that generation aligns with user intent.

\subsection{Aligning Context Inconsistency} Context inconsistency hallucination occurs when an LLM generates responses that contradict provided references, leading to factual misalignment. This often happens when segments of the response are not properly linked to relevant references. Attribution-based solutions address this by retrieving and integrating references to ensure consistency. This section examines two common techniques for mitigating context inconsistency hallucination: Pre-generation Prompt Engineering (T3) and Post-generation Response Correction (T4), along with their advantages and limitations.

\textbf{Pre-generation Prompt Engineering (T3)} Context inconsistency hallucination occurs when the model misaligns references with response segments, resulting in contradictions. Pre-generation prompt engineering methods use templates to guide the model in maintaining consistency with references \cite{In-contextRALM, QUIP, RECITE, Smartbook, PKG}. For example, In-Context RALM \cite{In-contextRALM} enhances alignment by concatenating the query with retrieved documents. QUIP \cite{QUIP} prompts the model to attribute claims explicitly. RECITE \cite{RECITE} encourages the model to "recite" relevant information before generating a response. SmartBook \cite{Smartbook} utilizes sub-questions and keywords for explicit citation, while PKG \cite{PKG} employs an open-source LLM to provide domain-relevant background information, enriching prompts for more concise and consistent responses.

\begin{table*}[h]
    \centering
    \footnotesize
    \renewcommand{\arraystretch}{1.5}
    \setlength{\tabcolsep}{4pt}
    \scalebox{0.8}{
    \begin{tabular}{>{\centering\arraybackslash}m{2cm}|>{\raggedright\arraybackslash}m{3cm}|>{\centering\arraybackslash}m{1.2cm}|>{\centering\arraybackslash}m{1.3cm}|>{\raggedright\arraybackslash}m{9.2cm}|>
    {\centering\arraybackslash}m{1.4cm}}
    \hline
    \rowcolor{gray!30}\textbf{Hallucination Type} & \multicolumn{1}{c|}{\textbf{Definition}} & \textbf{Retriever} & \textbf{Generator} & \multicolumn{1}{c|}{\textbf{Example}} & \textbf{Techniques} \\ \hline

    \textbf{Overconfidence} & 
    The LLM's response presents uncertain or nuanced information as absolute fact. &
     & \checkmark & 
Query: How to lose weight? \newline
Reference: [1] exercise. [2] \textcolor{blue}{diet, metabolism, and medical conditions}.  \newline
Response: The \textcolor{red}{only} way to lose weight is through exercise.
     &  T3, T4 \\ \hline

    \textbf{Outdatedness} & 
    The LLM's response was once correct but has become obsolete due to the passage of time. &
    \checkmark &  & 
    Query: Who is the current President of the United States? \newline
    Reference: \textcolor{blue}{Donald Trump} is now president (\textcolor{blue}{pre-2025 data}). \newline
    Response: \textcolor{red}{Joe Biden} is the current President of the United States.
     &  T1, T2 \\ \hline

    \textbf{Unverifiability} & 
    The LLM's response cannot be substantiated by existing knowledge, records, or sources. &
    \checkmark & \checkmark & 
    Query: What is the biological consequence of constructing the Eiffel Tower? \newline
    Reference: \textcolor{blue}{No historical or biological records}. \newline
    Response: The Eiffel Tower's construction led to the \textcolor{red}{extinction of the Parisian tiger}.
    &  T2, T4 \\ \hline

    \textbf{Instruction Deviation} & 
    The LLM's response fails to follow the user's explicit instructions. &
    \checkmark & \checkmark & 
    Query: \textcolor{red}{Translate} this sentence into Spanish: "The weather is nice today." \newline
    Reference: The sentence "The weather is nice today" is \textcolor{blue}{commonly understood by Spanish speakers} to imply pleasant weather conditions. \newline
    Response: The phrase "The weather is nice today" in Spanish conveys the idea of \textcolor{red}{good weather conditions}.
     &  T1, T2, T3 \\ \hline

    \textbf{Context Inconsistency} & 
    The LLM’s response contradicts or overlooks the retrieved references. &
     & \checkmark & 
    Query: Where is the Nile?  \newline
     Reference: The Nile’s source is in the Great Lakes region of \textcolor{blue}{central Africa}. \newline
     Response: The Nile originates in \textcolor{red}{Egypt}.
     &  T3, T4 \\ \hline

    \textbf{Reasoning Deficiency} & 
    The LLM’s response contains logical contradictions, invalid inference, or flawed reasoning chains. & 
     & \checkmark & 
    Query: Why do whales need to come to the surface? \newline
    Reference: Whales are mammals and breathe air through lungs, so \textcolor{blue}{they must surface regularly to inhale oxygen}. \newline
     Response: Whales live in the ocean, using gills to extract oxygen from water like fish. Therefore, \textcolor{red}{they don’t need to surface often}.
     &  T3, T4 \\ \hline

    \end{tabular}
    }
    \caption{Taxonomy of Hallucination Types in LLMs. This table categorizes various types of hallucinations generated by Large Language Models (LLMs), outlining their definitions, examples, the stage (retriever or generator) at which they occur and techniques can be used for mitigation.}
    \label{tab:hallucinations}
\end{table*}

\textbf{Post-generation Response Correction (T4)}
Even with correctly retrieved references, the generator may struggle to align responses, leading to context inconsistency hallucination. This can be mitigated through Post-generation Response Correction methods that compare responses with retrieved references to detect misalignments using an auxiliary model \cite{SourceCheckup, WebCiteS, EFEC} or the same LLM \cite{RSEGQA}. Mainstream methods use auxiliary models, like NLI models, to evaluate alignment. SourceCheckup and WebCiteS employ NLI-based verification at the statement level, flagging inconsistencies for correction. EFEC identifies key tokens in responses with an independent LLM, masking them, and instructing the LLM to fill in gaps based on retrieved references. Alternatively, RSEGQA prompts the same LLM to verify and revise its output by decomposing responses into sub-statements and assessing them against supporting references, editing contradictions using the retrieved evidence to ensure consistency.

\subsection{Reducing Reasoning Deficiency}
Reasoning deficiency hallucination often occurs in Chain-of-Thought (CoT) scenarios, where an LLM generates flawed or incomplete reasoning steps, leading to incorrect conclusions. Attribution-based solutions address this by retrieving relevant references for each step, using external evidence to enhance logical coherence and factual accuracy. This section explores two common attribution techniques to mitigate reasoning deficiency hallucination: Pre-generation Prompt Engineering (T3) and Post-generation Response Correction (T4), highlighting their strengths and limitations.

\textbf{Pre-generation Prompt Engineering (T3)} 
Reasoning deficiency hallucination can be mitigated through Pre-generation Prompt Engineering (T3) by structuring prompt templates that guide the model's reasoning process. These methods promote interpretable and logically sound step-by-step reasoning \cite{SubgraphRAG, Self-Reasoning}. For instance, SubgraphRAG \cite{SubgraphRAG} encodes references into a knowledge graph, identifying relevant subgraphs based on the query. These subgraphs are formatted into a prompt template, guiding the model to generate multi-step explanations grounded in structured evidence. Similarly, Self-Reasoning \cite{Self-Reasoning} develops a prompt template for each reasoning step, prompting the LLM to validate document relevance, extract key facts, and construct a coherent reasoning trajectory.

\textbf{Post-generation Response Correction (T4)} Post-generation Response Correction methods analyze and revise reasoning trajectories after generation, encouraging models to break down responses into explicit steps, assess coherence, and correct flawed logic \cite{Searchain, CoVe, Verify-and-edit}. Recent techniques employ various strategies for verification and correction. SearChain \cite{Searchain} uses predefined prompts to explicitly instruct the LLM to assess and regenerate flawed reasoning chains. In contrast, Verify-and-Edit \cite{Verify-and-edit} focuses on evaluating confidence scores for each step, flagging uncertainties, and generating retrieval-friendly questions to fetch evidence for corrections. CoVe \cite{CoVe} executes dedicated verification queries to validate claims against retrieved references, ensuring that only verified claims contribute to a revised, logically consistent response.

\subsection{Hallucination Handling using Attribution-based Techniques}
Attribution-based solutions can effectively address different types of hallucination by identifying their source: retriever or generator as shown in Table \ref{tab:hallucinations}. Retriever-related hallucinations arise from outdated, irrelevant, or incomplete references, often linked to issues like outdatedness and unverifiability. In contrast, generator-related hallucinations result from the model's response formulation, including overconfidence and reasoning deficiencies. This distinction is crucial for selecting appropriate techniques, such as Pre-retrieval Query Refining (T1), Post-retrieval Reference Identification (T2), Pre-generation Prompt Engineering (T3), and Post-generation Response Correction (T4).

\textbf{Retriever-oriented Techniques} focus on enhancing the quality, relevance, and timeliness of evidence retrieved before generation, using Pre-retrieval Query Refining (T1) and Post-retrieval Reference Identification (T2). T1 addresses hallucinations like outdatedness and instruction deviation by reformulating user queries for better alignment. Techniques include breaking complex queries into structured sub-queries, clarifying ambiguous inputs, and adding time-sensitive keywords, which help fetch more relevant and up-to-date references. T2 tackles outdatedness, unverifiability, and instruction deviation by filtering or re-ranking references based on quality and task alignment. Techniques include optimizing retrievers with auxiliary modules or reinforcement learning, using ranking models for semantic similarity, and maintaining updated knowledge bases. Some methods also refine attribution granularity with CoT-guided reasoning, ensuring only the most relevant and verifiable references are passed to the generator, reducing hallucination risks.

\textbf{Generator-oriented Techniques} aim to regulate hallucinations that occur during the response generation process, utilizing Pre-generation Prompt Engineering (T3) and Post-generation Response Correction (T4). T3 effectively addresses hallucinations such as overconfidence, context inconsistency, instruction deviation, and reasoning deficiency. This approach involves designing structured prompt templates with rule-based prompts, incorporating task-specific keywords, and employing structured reasoning formats. T3 also includes selecting from a pool of templates to guide the LLM in producing responses that are more accurate and aligned with the task. For example, prompt templates can incorporate hedging language to mitigate overconfidence, enforce citation formats for improved grounding, or scaffold responses into step-by-step reasoning to enhance logical coherence. T4 is applied after the response is generated and targets hallucinations like overconfidence, unverifiability, context inconsistency, and reasoning deficiency. This approach operates in two main ways: (1) verifying and correcting a single response, or (2) generating multiple candidate responses and selecting the most reliable one. For single-response correction, techniques include prompting the LLM to self-critique through verification or clarification questions, or using auxiliary models such as NLI classifiers, reward models, or fine-tuned LLMs to assess factual consistency. For multi-response selection, systems may employ ranking strategies based on confidence scores, factual alignment, or human preference-trained reward models (e.g., via RLHF). In both cases, the results of these evaluations are used to revise or regenerate flawed segments, ensuring that responses are more accurate, consistent, and verifiable.

\subsection{Discussion on Challenges}

While attribution-based techniques have demonstrated strong potential in mitigating various forms of hallucination, several limitations affect their reliability, generalizability, and practical deployment. These challenges highlight opportunities for future research and system refinement:

\textbf{Single-Source Validation.}
Many existing approaches evaluate the verifiability of model responses against a single source or a limited set of references. This may be insufficient for open-domain or multi-perspective questions, where claims require validation across multiple diverse sources. Relying on a narrow evidence base risks missing contradictions, overlooking alternative views, or reinforcing biased content.

\textbf{Reliance on External Databases and Safety Concerns.}
Attribution methods often depend on external databases (e.g., Wikipedia, web search) that may include incomplete, incorrect, or unsafe information. This raises concerns about factual accuracy, potential misinformation, and copyright violations. Furthermore, systems using real-time web access (e.g., WebGPT) face the added complexity of verifying the credibility and legality of dynamic sources, which can be volatile and unmoderated.

\textbf{Use of LLMs as Retrieval Judges.}
Recent approaches like LLatrieval and CoVe employ LLMs to verify retrieval quality or judge reasoning consistency. However, this introduces circularity: using the same class of models, which are prone to hallucination, as evaluators may inherit similar biases or overlook subtle reasoning errors. Moreover, there is limited consensus on how reliable LLMs are in scoring retrieval quality or verifying complex claims without external calibration.

\textbf{Limitations in Long-Context Understanding.}
Current LLMs and retrieval systems still struggle with long or multi-document contexts. Many approaches truncate input, leading to missing evidence, or suffer from degraded performance when asked to reason over large spans. This affects the alignment of context and the integrity of generated reasoning, particularly in knowledge-dense tasks like multi-hop QA or document-level summarization.

\textbf{Error Accumulation in CoT Reasoning.}
While CoT prompting improves interpretability and structure, it introduces the risk of hallucination propagation. An incorrect early step can mislead the subsequent reasoning path, especially in long chains of reasoning. Post-hoc correction methods help address this, but robust intermediate validation is still lacking in many systems.

\textbf{One-shot vs. Few-shot Prompting.}
Many current solutions rely on carefully engineered one-shot prompts or static templates. These are often brittle across domains or tasks, failing to generalize without significant tuning. Few-shot approaches with in-context demonstrations offer greater flexibility but come with increased token usage and may still struggle with unseen reasoning patterns or instruction deviations.

\section{Conclusion}
This survey explores attribution-based techniques aimed at reducing hallucinations in RAG-based QA systems. We propose a unified pipeline consisting of four key components: query refining, reference identification, prompt engineering, and response correction. Additionally, we categorize different types of hallucinations and their corresponding mitigation strategies, with the goal of enhancing factual consistency in LLMs.

\section{Limitations}
Despite the advancements in attribution-based techniques within Retrieval-Augmented Generation systems to mitigate hallucinations, several limitations persist. First, the complexity of implementing these techniques can hinder their practical application, leading to inconsistencies in performance and making it challenging for practitioners to integrate them into existing frameworks effectively. Additionally, there is a lack of comprehensive evaluation and systematic comparison of these techniques across different RAG systems. This absence of standardized assessment makes it difficult to identify the most suitable approach for specific applications, potentially limiting the broader adoption and optimization of RAG systems in various contexts.

\bibliography{tacl2021}
\bibliographystyle{acl_natbib}

\end{document}